\documentclass[11pt]{article}
\usepackage{geometry}                
\usepackage{graphicx}
\usepackage{amssymb}
\usepackage{epstopdf}
\usepackage{setspace}
\usepackage{natbib}
\usepackage{amsmath}

\newcommand{\vect}[1]{\boldsymbol{#1}}                 
\usepackage{listings}
\usepackage{url}
\usepackage{array}
\newcolumntype{H}{>{\setbox0=\hbox\bgroup}c<{\egroup}@{}}

\title{Maximum likelihood estimation of a finite mixture of logistic regression models in a continuous data stream}

\author{Prof. dr. M.C. Kaptein
  \and
  Dr. P. Ketelaar
}

\newcommand{\Addresses}{{
  \bigskip
  \footnotesize

  M.C.~Kaptein (Corresponding author), \textsc{Jheronimus Academy of Data Science}, Den Bosch, the Netherlands \and
\textsc{Statistics and Research Methods}, Tilburg University, Tilburg, the Netherlands\par\nopagebreak
  \textit{E-mail address}, M.C.~Kaptein \texttt{m.c.kaptein@uvt.nl}

  \medskip
  P.~Ketelaar , \textsc{Behavioural Science Institute (BSI)}, Communication, Radboud University, Nijmegen, the Netherlands.\
}}

\date{}                                           

\begin{document}
\maketitle

\section*{Abstract}
In marketing we are often confronted with a continuous stream of responses to marketing messages. Such streaming data provide invaluable information regarding message effectiveness and segmentation. However, streaming data are hard to analyze using conventional methods: their high volume and the fact that they are continuously augmented means that it takes considerable time to analyze them. We propose a method for estimating a finite mixture of logistic regression models which can be used to cluster customers based on a continuous stream of responses. This method, which we coin \emph{oFMLR}, allows segments to be identified in data streams or extremely large static datasets. Contrary toÒblack boxÓ algorithms, oFMLR provides model estimates that are directly interpretable. We first introduce oFMLR, explaining in passing general topics such as online estimation and the EM algorithm, making this paper a high level overview of possible methods of dealing with large data streams in marketing practice. Next, we discuss model convergence, identifiability, and relations to alternative, Bayesian,  methods; we also identify more general issues that arise from dealing with continuously augmented data sets. Finally, we introduce the oFMLR [R] package and evaluate the method by numerical simulation and by analyzing a large customer clickstream dataset. 

\medskip
\noindent
\emph{Keywords:} Clustering, Finite mixture models, Logistic regression, Online estimation, Expectation-Maximization.

\section{Introduction}
In marketing we are often confronted with continuous data streams of customer behavior \citep{Su2015, Chatterjee2003, Moe2004a, Bucklin2009}. Consider online advertising which has become a major advertising medium, with over $110$ billion USD invested in 2014 in search engines, social media, and general websites \citep{Yeu2013}. Online advertising is an extremely competitive medium, and customers attention is not easily won: Numerous studies have shown that customers deliberately avoid internet ads \citep{Yaveroglu2008, Cho2004, Rojas-Mendez2009, Baek2012, Seyedghorban2015}. As such, getting ads noticed and processed by users is of particular importance \citep{Huang2006, Yeu2013, Yaveroglu2008}. This is not easy, and marketing and advertising scholars have spent considerable effort studying online customer behavior in the last decades \citep[see, e.g.,][]{Ranaweera2005, Huang2006, Micheaux2011, Bleier2015, Vakratsas2004}. In these  studies, scholars have stressed the difficult challenge of understanding online customer behavior \citep{Yeu2013}. 

On a positive note, online customer behavior can potentially make it possible to measure the effectiveness of marketing messages, experiment with novel types of messages, and understand customer responses better \citep{Huang2006}. Since online behavioral responses of customers can often be measured directly on  scale hitherto unknown, online advertising and retailing opens up a treasure throve of information: we can now obtain direct, near real-time, data on customers. This information should allow us to develop and test new marketing theories  \citep{Laczniak2015}. 

However, to use these emerging opportunities for research and marketing practice, we need to have sound analysis methods capable of dealing with novel data sources. For example, while it is potentially extremely useful, analyzing click stream data is challenging for several reasons. First, click stream data sets grow very large: databases containing responses (clicks) to online advertisements quickly exceed millions of records, often calledÒrowsÓ. Since we often also have access to a large number of customer background characteristics (e.g., their search queries, their IP address, their geo-location, and their purchase behavior), the database often also contains a large number of ÒcolumnsÓ. The sheer volume of data poses problems. But not only does the volume poses problems: click stream data are often continuously augmented. New data points arrive continuously, and often at high velocity. Hence, to truly utilize the data-sources at our disposal, we need analysis methods that can deal with both high volume and high velocity data \citep{Boyd2012}.

Recently, many technologies and statistical methods have been proposed to deal with Big Data or high velocity data. On the one hand, a large number of technical advances have been made: infrastructures such as Hadoop, Apache Spark, and the Map/Reduce framework allow us to store and process larges volumes of data \citep{Chu2007, Alpcan2009}. These software packages are, however, geared towards computer scientists, not marketing scholars. On the other hand, scholars in statistics and machine learning are actively working on novel estimation methods to deal with high volume and high velocity data. Examples include methods for fitting generalized linear models in data streams using gradient descent methods \citep{Zinkevich2010}, as well as computationally efficient bootstrap methods \citep{Owen2012}.

This paper proposes another estimation method that is well suited for the analysis of click-stream responses to marketing messages: we present an \emph{online} (or \emph{streaming}) method for estimating Finite Mixtures of Logistic Regression models which we coin \emph{oFMLR}.
Our method will appeal to marketing scholars for a number of reasons: first, it is well suited for analyzing the dichotomous response data that are often observed in online advertising and retailing. Furthermore, the model is capable of dealing with extremely high volume and high velocity data, thus making it a perfect candidate for analyzing click stream data. Finally, it builds on the well-known logistic regression model that is frequently used in marketing research \citep[see, e.g.,][for examples of the use of logistic regression]{Yeu2013, Bonfrer2009, Kireyev2015}. This has the advantage that marketing scholars can readily interpret the estimated model parameters unlike the ``black box'' models often used in machine learning.

The proposed oFMLR is related to recent attempts to cluster customers in large datasets: for example, \citet{Su2015} proposed a specific method of deriving customer interest segments from behavioral observations in online commerce. Our method is, however, more general in the sense that it does not rely on particularities of internet usage such as the frequency of customer visits: our method can be used to uncover segments in data streams, as long as the primary outcome measure is dichotomous. OFMLR is a specific implementation of the online expectation-maximization (EM) algorithm more generally described by \citet{Cappe2009} and allows clustering (or segmentation) of customers during a data-stream: the model allows one to identify homogenous groups of customers without using ad-hoc analysis.

Mixture models, more specifically mixtures of logistic regression models, have been used frequently and effectively in the marketing literature for decades. Already two decades ago, \citet{wedel1993latent} discussed latent class Poisson regression models and more general mixture models \citep{wedel1995mixture}, in work that remains influential in marketing practice today. Applications of finite mixtures of regression models date even further back, with \citet{kamakura1989probabilistic} applying mixture models for customer segmentation. Research into the use of mixture models for applied market segmentation is still thriving \citep{wedel2012market}, and, despite having been known for decades, their use is still increasing. Mixture models are an invaluable tool for analyzing marketing data and identifying consumer segments. Unfortunately, many estimation methods of mixture models are computationally demanding for extremely large or continuously augmented datasets. Our aim is to solve this problem for a specific class of models. While presenting our solution, we also emphasize the issues that arise when analyzing high velocity data streams more generally, and discuss several possible approaches.

The next section of this paper presents some of the conceptual building blocks of our proposed method. First, we introduce the notation that is used. Second, we discuss possible methods of dealing with high volume or high velocity data, focussing explicitly on online estimation as a preferred approach. Third, we discuss estimation of a simple logistic regression model in a data stream using stochastic gradient descent (SGD). Next, we detail oFMLR: we present an algorithm for estimating a finite mixture of logistic regression models in a data stream. After formalizing the model and discussing the Expectation-Maximization (EM) algorithm as a general method for estimating latent variable models, we present oFMLR and discuss how the estimated parameters can be interpreted. In order to explain the issue to a broader audience, we introduce the methods for dealing with high velocity data, SGD, and EM quite broadly. Next, we further discuss convergence, identifiability, and model selection, and we evaluate oFMLR using a simulation study and by analyzing an actual dataset. Finally, we show the practical use of oFMLR and discuss possibilities for future work.

\section{Big Data Analysis in Online Marketing}

This section presents several conceptual solutions for dealing with extremely high volume or high velocity data streams, and introduces an estimation method for logistic regression models\footnote{Note that the same method is more broadly applicable for maximum likelihood estimation---our example however focusses on logistic regression.} that is well-suited for dealing with high volume and velocity data.

Before discussing the possible methods of dealing with large datasets, we first introduce our notation. We assume that the data of interest consists of clicks on marketing messages (e.g., advertisements or product displays) that arrive \emph{sequentially} over time. We further assume the dataset is continuously augmented. We will denote $y_t \in \{0, 1\}$ for a click on a message at a specific point in time $t=1, \dots, t=T$ (where the horizon $T$ might not be known). We will assume that clicks on messages originate from different customers, and hence by analogy the subscript $t$, would often, in the analysis of static datasets be referred to using $i=1, \dots, i=n$ where $n$ is the total number of responses; we will use $y_i$ explicitly when discussing static analysis methods. Next to our dependent variable of interest $y$, we often know a number of properties of the message displayed (e.g., the content of the advertisement), and we know a number of customer characteristics as well. We will denote this information using the feature vector $\vect{x}_t$. 

When considering the analysis of a static dataset we will sometimes refer to vector $\vect{y} = (y_1, \dots, y_n)$, and the ($n \times p$) matrix $\vect{X}$ as the data. Here, $p$ is the total number of features (often including a column of 1's for the intercept of a model). The full dataset $\mathcal{D}$, is thus composed of rows $\vect{d}_1 = (y_1, \vect{x}_1), \dots, \vect{d}_T= (y_T, \vect{x}_T)$. Finally, we will use $\vect{\theta}$ quite generally to denote the quantities of interest to a researcher (e.g., the parameters of some model), and $\hat{\vect{\theta}}_t$ for the estimated state of the parameters at time $t$.

\subsection{Methods of Dealing with ``Big Data''}
\label{sub:methods}

Analyzing the click-stream data resulting from online marketing is frequently a Big Data problem for a variety of  reasons: first, the number of observations $T$ is large. This might be cumbersome since analysis using standard statistical software may take too long. Second, the data are often continuously augmented; thus, $T$ grows continuously. This is cumbersome not just when $T$ is large, but also when the data arrive at large speeds: It is not feasible to re-analyze the newly resulting dataset every time the old dataset is augmented.\footnote{The dimensionality of the feature vector $\vect{x}_t$ might also pose problems in the analysis. Many feature selection and dimensionality reduction methods have been developed to deal with such problems \citep[see, e.g.,][]{Yue2012, Gaber2005}. These are outside the scope of this paper.} There are, however, several methods available to deal with large, or quickly augmented datasets. Broadly, the following four approaches can be identified:

\begin{itemize}
\item \emph{Subsampling:} A simple method of dealing with a dataset that is too large is subsampling from the collected data. Here we randomly  select $m < T$ rows from the dataset $\mathcal{D}$, without replacement. Clearly, if $m$ is chosen such that the resulting dataset can be dealt with using conventional software, the problem is solved; it is, however, solved at the cost of losing the information from data points not included in the sample. Subsampling can also be done in data streams: in this case, one randomly ignores a proportion of the incoming datapoints.
\item \emph{Moving Window:} : A second approach for dealing with large datasets is to subsample only the last $m$ elements of the data and use those for analysis. Thus, at each time point, the analyzed data are composed of rows $\vect{d}_{T-m},\dots, \vect{d}_T$. If $m$ is chosen such that the analysis can be carried out faster then the dataset is augmented, the problem is solved; once again, however, it is solved at the cost of ignoring available, historical, information.
\item \emph{Parallelize:} A method that has become popular over the last decade is parallelization. Effectively, if one computing machine (often referred to as \emph{core}) cannot deal with all the data, then we split the data over $c$ cores. Each core receives its own batch of $T/c$ rows, performs the analysis on its batch, and the results are combined. This primarily solves the problem of analysis becoming too lengthy; the computation time is effectively used in parallel as opposed to in sequence, and is thus often greatly reduced. Note, however, that not all analysis methods can be easily parallelized: some methods that are relatively standard in advertising, such as mixture models \citep{Cheong2011}, are not easily expressed in separate, independent batches \citep[for more on this topic see][]{Poggio2011}
\item \emph{Online learning (or streaming):} A fourth possible approach is called ``online learning'', which is a somewhat confusing name in situations where the data is collected online (the latter referring to Òon the WebÓ, the former referring to the estimation method). The basic idea behind online learning is to identify the parameters $\vect{\theta}$ of interest up front, and subsequently to make use of an \emph{update} function of the parameters each time a datapoint enters: $\vect{\theta}_{t} = f(\vect{\theta}_{t-1}, \vect{d}_{t})$. 
\end{itemize}

In this paper we will focus solely on online learning since a) it is based on all the datapoints (as opposed to subsampling or moving window approaches), and b) it is extremely well-suited for dealing with high velocity data streams: as long as the update of $\vect{\theta}$ can be computed quickly, online learning can provide accurate estimates at each point in time. Note that we will sometimes use the shorthand notation $\vect{\theta} := f(\vect{\theta}, \vect{d})$ to denote the parameter update as a function of the old parameter and a single datapoint.

To illustrate the computational advantage of online estimation, consider the computation of a simple sum: we merely want to know the number of views of an advertisement. Hence, we maintain a counter $\theta^{count}$ denoting the number of views. Conventional analysis would compute this as follows: $\theta^{count}_T = \sum_{t=1}^T 1$. Each time a new datapoint enters, this count would have to be computed again: $\theta^{count}_{T+1} = \sum_{t=1}^{T+1} 1$. It is easily seen that this requires $1 + 2 + \dots + T-1+T = \frac{1}{2} T(T+1)$ computations. Using online learning one would specify: $\theta^{count}_{T+1} = \theta^{count}_T + 1$ (or simply $\theta := \theta+1$); a very simple \emph{update} function, using just \emph{one} computation for each new datapoint and thus $T$ computations in total ($1 + 1 + \dots$).  This tremendous difference in the computational complexity of computing a count offline vs. online clearly highlights the appeal of online methods for fitting statistical models in high volume or high velocity data streams. Luckily, more than just sums can be computed exactly online: means, variances, covariances, linear models, and a number of other interesting quantities can be computed using an update function rather than by revisiting all historical data \citep{ippel2017dealing}. However, exact online estimation methods are not available for more complex models---such as the mixture model we discuss below. The next section discusses possible approximation methods to fit more complex models. Note that we focus on frequentist estimation; alternative Bayesian methods will be discussed in Section \ref{sec:Bayes}.

\subsection{Online Learning of Logistic Regression}

This section discusses how a logistic regression model can be estimated using online learning. We aim to model the observations $y_i$ which are a Binomial random variable whose proportion parameter $p$ depends on $\vect{x}_i$. Hence, $y_i \sim \mbox{bin}(n_i, p(\vect{x}_i))$ using a logit link:
\begin{equation}
p(\vect{x}_i) = \frac{1}{1 + e^{- \vect{x}_i \vect{\beta}} }
\end{equation}
where $\vect{x}_i \vect{\beta}$ is called the \emph{linear} part of the model and the parameter vector $\vect{\beta}$ of the model is easy to interpret \citep[for more details see][]{Huang2006, Micheaux2011, Yaveroglu2008}. 

In a conventional frequentist offline analysis, the parameters  are generally estimated using maximum likelihood. The likelihood of the data as a function of the parameters is given by
\begin{eqnarray}
\begin{split}
L(\vect{\beta} | \mathcal{D} ) 	& =  \prod_{i=1}^{n} p(\vect{x}_i)^{y_i}(1-p(\vect{x}_i))^{1-y_i} \\
						& =  \prod_{i=1}^{n} \left( \frac{1}{1 + e^{- \vect{x}_i \vect{\beta}} } \right)^{y_i}  \left( 1-\frac{1}{1 + e^{- \vect{x}_i \vect{\beta}} } \right) ^{1-y_i}
\end{split}
\label{fun:likelyhood}
\end{eqnarray}
For computational convenience, one often takes the log of the likelihood (which simplifies the the computation by effectively replacing the product term by a sum), and subsequently finds the maximum of the log-likelihood $l(\vect{\beta} | \mathcal{D} )$ as a function of $\vect{\beta}$ by setting its derivative to $0$. As usual, we will denote the resulting estimate $\hat{\vect{\beta}}$. However, in the case of logistic regression, this procedure results in a system of nonlinear equations that is not easily solved; the solution cannot be derived algebraically and must be numerically approximated.

However, other methods are available, a particularly simple one being Gradient Descent (GD), or Ascent (GA) in the case of maximization: with GD we compute the gradient---the vector of first order derivatives---of the log-likelihood, $\nabla l(\vect{\beta} | \mathcal{D} )$, and evaluate it at some initial value of the parameters $\hat{\vect{\beta}}^{(0)}$. Next, we apply the following iterative scheme:
\begin{eqnarray}  
\hat{\vect{\beta}}^{(i+1)}= \hat{\vect{\beta}}^{(i)} + \lambda \nabla l(\hat{\vect{\beta}}^{(i)} | \mathcal{D} ).
\end{eqnarray}
Hence, we  make a step at each iteration (the size of which is determined by $\lambda$) in the direction of the gradient.

This procedure is intuitive: consider finding the maximum of a parabola, $f(x) = -(x-a)^2$ where $a$ is some constant (and the sought-for value of $x$ that maximizes $f(x)$). If, $f'(x) > 0$, (the derivative evaluated at $x$ being positive) the curve is going up, and thus the maximum $a$ must be to the right of $x$. Thus, one has to make a step towards higher values of $x$. Similarly, if $f'(x) < 0$, the curve is sloping down, and lower values of $x$ would approach the maximum. Gradient Ascent formalizes this intuition, and generalizes it for multidimensional cases. Given an appropriate step size $\lambda$, GA will converge to the maximum likelihood estimate \citep{Poggio2011}. 

It is relatively simple to transform parameter estimation using GD into an online learning method called \emph{Stochastic} Gradient Descent (SGD); we merely change from updating our parameter estimates $\vect{\beta}$ using multiple iterations through entire dataset $\mathcal{D}$ to updating our estimates for each new data point $D_t$: 
\begin{eqnarray}  
\hat{\vect{\beta}}_{t+1} = \hat{\vect{\beta}}_{t} + \lambda \nabla l(\hat{\vect{\beta}}_t | \vect{d}_t ),
\end{eqnarray}
or equivalently, 
\begin{eqnarray}  
\hat{\vect{\beta}} := \hat{\vect{\beta}} + \lambda \nabla l(\hat{\vect{\beta}} | \vect{d} )
\end{eqnarray}
For specific models and specific learn rates, one can show that SGD converges to the maximum likelihood estimate \citep{Poggio2011}. 
). Logistic regression is especially well behaved in this case, but for an extensive discussion of SGD learning---and convergence---rates we refer the reader to \citet{toulis2014statistical}.
Note that SGD is a useful online estimation method for any model for which the derivative of the log-likelihood is easy to evaluate: hence, SGD provides a general method for fitting statistical models in data streams  \citep{Zinkevich2010}. However, if the gradient of the log-likelihood is not easy to evaluate, then SGD might be cumbersome.

\section{Mixed Logistic Regression Models in Data streams}
\label{mix:log:online}

Logistic regression models are frequently used to analyze click-stream data. In the previous section, we have illustrated how this model can be estimated in a high volume, high velocity data stream using SGD. However, marketing click-stream data are often generated by diverse groups of customers. While the logistic regression model is well-suited to deal with such heterogeneity if these clusters (or segments) of customers are known, this is not always the case: though customers can often be meaningfully grouped into relatively homogenous clusters, the cluster memberships are not known a priori. Rather, the \emph{latent}, unobserved, cluster memberships can only be derived ad hoc from the collected data. Such clustering is quite often undertaken in advertising and marketing research \citep[e.g.,][]{Bakshy2012, McFarland2006, Su2015, Jedidi1997, Kireyev2015} and is meaningful for marketing theory: identifying clusters of consumers and interpreting model estimates can greatly improve our understanding of the effects of marketing messages.

\subsection{A Finite Mixture of Logistic Regression Models}
The finite mixture of logistic regression models offers a suitable clustering method for online advertising \citep{Gortmaker1994, Wang1998}. This model extends the logistic regression model by assuming that the observed data originate from a number of unobservable clusters $K$ that are characterized by different regression models. Effectively, one assumes that there are $k=1,\dots,k=K$ homogenous clusters of customers, each with its own relationship between the feature vector $\vect{x}$ and the observed clicks $y$. For example, one could imagine two groups of customers differing in their responses to product pricing: One group of customers ($k=1$) might be seeking exclusivity, and thus attracted by high prices, while another group $k=2$ is looking for bargains. This means that a positive slope of price would be expected for the first group, all else being equal, while a negative slope would be expected for the second group.

An intuitive way of thinking about mixture models is to simply extend the logistic regression model presented in Equation \ref{fun:likelyhood}. We can denote the likelihood of the finite mixture model by:
\begin{eqnarray}  
L(\vect{\theta} | \mathcal{D}) = \prod_{i=1}^{n}  \sum_{k=1}^K \alpha_k \left( \frac{1}{1 + e^{- \vect{x}_i \vect{\beta}_k} } \right)^{y_i}  \left( 1-\frac{1}{1 + e^{- \vect{x}_i \vect{\beta}_k} } \right) ^{1-y_i}
\label{likelihood:ofmlr}
\end{eqnarray}
\noindent
where the proportion parameter $p(x_i)$ of the observed $y_i$'s is modeled by a separate vector of coefficients $\vect{\beta}_k$ within each unobserved cluster. The cluster membership probabilities $\alpha_k$ ($\sum_{k=1}^K \alpha_k = 1$) are considered ``mixing probabilities'' which relate to the size of the clusters. 

One could interpret the model as originating from the following sampling scheme: first, a cluster $k$ is drawn from a multinomial distribution with probabilities $\alpha_1, \dots, \alpha_k$. Second, \emph{given} the cluster, the observation $y_i$ is generated from the corresponding model with parameters $\vect{\beta}_k$. Note that we will often use $\vect{\theta}$ to refer to both the mixture probabilities ($\vect{\alpha}$) as well as the regression coefficients within each cluster ($\vect{\beta}_1, \dots, \vect{\beta}_k$).

\subsection{Estimating Mixture models: The EM algorithm}

Estimating the finite mixture model is more involved  than estimating simple regression models since the log-likelihood of the data is not merely a summation over datapoints, but rather a summation over data points and often unknown clusters \citep{Wang1998}. This complicates estimation as the gradient of the (log-)likelihood is not easily evaluated. This complicates estimation, as the gradient of the (log-)likelihood is not easily evaluated. However, \emph{if the cluster membership of each customer were known}, then the estimation is simple: if we know in advance to which cluster a customer belongs, then we can simply estimate separate logistic regression models. This occurs frequently: for many latent variable or latent class models, estimation is greatly simplified if the class memberships would have been known. The EM algorithm is frequently used to estimate parameters in such situations.

To give an idea to how the EM algorithm tackles the unobserved data problem, we consider a relatively simple example \citep[inspired by the explanation in][]{Do2008}. Consider observing a total of $50$ coin tosses originating from the tossing of two coins, A and B, in batches of ten. Hence, the data could be fully described using two vectors, each with five elements, where the vector $\vect{y}=(y_1,y_2,\dots,y_5)$ denotes the number of heads in each of the five batches (thus $y_i \in \{0,\dots,10\})$, and $\vect{z}=(z_1, z_2, \dots, z_5)$ denotes the coin that was tossed $z_i \in \{A, B\}$. Our objective is to estimate the probability that each coin will land heads, which we will denote $\vect{\theta}=(\theta_A, \theta_B)$. A complete observed dataset could be: $\vect{y} = (3,8,2,9,7)$ and $\vect{z} = (A, B, A, B, B)$ which intuitively raises the idea that $\theta_B > \theta_A$.

Given the \emph{full} dataset, maximum likelihood estimation is simple:
\begin{eqnarray}
\hat{\theta_A} 	& = \frac{ \text{\# of heads using coin A} }{ \text{\# of tosses using coin A} }
			& = \frac{ 3 + 2 }{ 10 + 10 } = .25
\end{eqnarray}
and similarly for $\hat{\theta_B} = \frac{24}{30} = .8$. This is analogues to knowing the cluster $k$ for our finite mixture model, and merely using a conventional ML approach within each cluster. However, what can we do if we \emph{only} observe the vector $y$, and not $z$? 

Formally, we are seeking a method to maximize the log-likelihood of the observed data $l(y; \vect{\theta})$. The problem is easy when we observe all the data and we are looking for the \emph{complete} data solution, $l(y, z; \vect{\theta})$, as demonstrated above. The EM algorithm allows us to solve the \emph{incomplete} data case, where $z$ is not observed. The iterative algorithm starts with a ``guess'' of the parameters. For example we can set $\hat{\theta}_A = .4$ and $\hat{\theta}_B = .5$. Next, we iterate the following steps:
\begin{enumerate}
\item \emph{Expectation (E) step}: In the expectation step, a probability distribution over the missing data is ``guessed'' given the current model parameters. In the coin tossing example, for each batch we compute the probability of that batch originating from coin A or coin B. For example, for batch 1 (3/10), and $\hat{\theta}$ as specified above, the probability of the data originating from coin A is $.65$, while it is $.35$ for B. This is intuitive: given that our initial estimate of $\hat{\theta}_A$ is closer to $3/10 = .3$ then our initial estimate $\hat{\theta}_B$ the data in batch one is more probable for coin A then for coin B.\footnote{These probabilities are computed by evaluating the binomial density  $f(y_1;n=10,\theta) = \Pr(Y = y_1) = \binom{10}{y_1} \hat{\theta}^y_1(1-\hat{\theta})^{10-y}$ for both $\hat{\theta}_A$ and $\hat{\theta}_B$ and normalizing. This results in vectors $\vect{p}_a = (0.65, 0.19, 0.73, 0.14, 0.27)$ and $\vect{p}_b = (0.35, 0.81, 0.27, 0.86, 0.73)$ respectively for the probability that coin A or coin B generated the data in the respective batch.}
\item \emph{Maximization (M) step}: In the maximization step, the model parameters are updated using the guessed probability distribution over the missing data. Here, we compute the maximum likelihood estimate as if we observed z, but, we weight the data according to the probability that it originates from one of the coins. Hence, we are not certain which cluster it belongs to (either A or B as denoted in vector $z$) but rather assign it probabilistically. This gives us the new estimates $\hat{\theta}_A$ and $\hat{\theta}_B$ which are used in the next expectation step.\footnote{The updated $\hat{\theta}_A$ is computed by multiplying $\vect{p}_a$ by $\vect{y}$ to obtain the ``weighted'' number of successes, which gives $(1.9, 1.6, 1.5, 1.2, 1.9)$. The same is done for the failures, and based on these weighted observations the guesses of the parameters are updated; this leads to $\hat{\theta}_A = .408$ and $\hat{\theta}_B = .693$ after the first iteration.}
\end{enumerate}
The numerical values of the estimate $\hat{\theta}_A$ in subsequent iterations are $.40, .41, .30, .26$, and $.25$ respectively; hence, the algorithm seems to have converged in as few as five iterations.

The EM algorithm is very flexible, and can be used for a large number of problems that can be formulated as a missing data problem. We refer the interested reader to \citet{Do2008} for a general introduction, and to \citet{Moon1996} or \citet{Dempster1977} for a thorough mathematical overview. The algorithm is well-known, and the EM-solution for the offline estimation of a mixture of logistic regression models is also well-known \citep{Wang1998}. The EM algorithm for a mixture of logistic regression models can be understood as follows:
\begin{enumerate}
\item \emph{Expectation (E) step}: Given initial values of $\hat{\vect{\alpha}}$ (the mixture probabilities) and $\hat{\vect{\beta}}_k$ (the regression coefficients in each cluster), we compute the expected cluster membership probability of each observation, $z_{ik}$ (thus, we compute a ``probabilistic assignment'' to $k$ for each observed unit $i$). 
\item \emph{Maximization (M) step}: Given the conditional cluster membership assignments as computed in the E step, we fit a \emph{weighted} logistic regression (using conventional ML estimation) for each cluster $k$ to obtain an update of $\hat{\vect{\beta}}_k$. The update of $\hat{\vect{\alpha}}$ is obtained by averaging over the expected membership probabilities: $\hat{\alpha}_k = \sum_{i=1}^n z_{ik} / n$
\end{enumerate}
This iterative scheme, when combined with SGD, is surprisingly easy to implement online, as will be demonstrated in the next section.

\subsection{Online Finite Mixture of Logistic Regressions}

Here we present our method for fitting the oFMLR model in full. The algorithm starts with a choice of $K$ (fixed---see Section \ref{sec:choosek} for a discussion), and starting values for $\hat{\vect{\alpha}}$ and each $\hat{\vect{\beta}}_1,\dots,\hat{\vect{\beta}}_K$ (which we jointly refer to as $\hat{\vect{\theta}}$). Next, for each arriving datapoint $\vect{d}_t$ we compute the following:
\begin{enumerate}
\item \emph{Expectation (E) step}: In the expectation step the conditional probability of the datapoint $\vect{d}_t$ beloning to a cluster $k$ is computed given the current state of $\hat{\vect{\alpha}}$ and $\hat{\vect{\beta}}_k$. Thus, we compute
\begin{eqnarray}
z_{tk} & = & \frac{ \hat{\alpha}_k p(\vect{x}_t)^{y_t}(1-p(\vect{x}_t))^{1-y_t} }{ \sum_{k=1}^K \hat{\alpha}_k p(\vect{x}_t)^{y_t}(1-p(\vect{x}_t))^{1-y_t} }
\end{eqnarray} 
where $p(\vect{x}_t) = \frac{1}{1 + e^{- \vect{x}_t \hat{\vect{\beta}}_k} }$ for each $k$. Note that $z_{tk} = f(\hat{\vect{\theta}}, \vect{d}_t)$ and is thus an online update.
\item{ \emph{Maximization (M) step}: In the M step we update our estimates of $\hat{\vect{\beta}}_k$ and $\hat{\vect{\alpha}}$:\footnote{Note that we use $\hat{\alpha}_k$ below, where the $k$ refers to the different elements of $\hat{\alpha}$, as opposed to $\hat{\vect{\beta}}_k$ where each $k$ denotes a different vector $\hat{\vect{\beta}}$ for the respective model.}}
\begin{itemize}
\item We first update each $\hat{\vect{\beta}}_k$ online using SGD weighted by $z_{tk}$:
\begin{eqnarray}
\hat{\vect{\beta}}_k 	& := &  \hat{\vect{\beta}}_k + \gamma \nabla l (\hat{\vect{\beta}}_k | \vect{d}_t, z_{tk}) \\
				& := &  \hat{\vect{\beta}}_k + \gamma z_{ik} (y_t - p(\vect{x}_t)) \vect{x}_t
\end{eqnarray}
where again $p(\vect{x}_t) = \frac{1}{1 + e^{- \vect{x}_t \hat{\vect{\beta}}_k} }$. 
\item We then update $\hat{\alpha}$ by computing an online average of $z_{tk}$:
\begin{eqnarray}
\hat{\alpha}_k	& := & \hat{\alpha}_k + \frac{ z_{tk}-\hat{\alpha}_k }{t}
\end{eqnarray}
\end{itemize}
\end{enumerate}
We thus maintain in memory a counter $t$, a $k$-dimensional vector $\hat{\vect{\alpha}}$ denoting the mixing probabilities, and $k$ $p$ dimensional vectors $\hat{\vect{\beta}}_k$, each of which is updated fully online. An efficient [R] S4 package to fit the above model can be downloaded and installed from \url{https://github.com/MKaptein/ofmlr} (the package is discussed in more detail in Section \ref{sec:empirical}).

\subsection{Identifiability of Model Parameters}

Mixture models are appealing for their ease of interpretation and intuitive representation as a latent variable model. However, mixture models also provide challenges; we have already mentioned the difficulty of estimating the parameters of the mixture model. This section discusses another challenge: identifiability. A given mixture model is said to be identifiable if it is uniquely characterized, so that no two distinct sets of parameters defining the mixture yield the same distribution. Identifiability of distinct mixture models has been actively studied: \citet{teicher1963identifiability} and \citet{teicher1967identifiability} provided conditions under which mixtures of binomial distributions are identifiable. \citet{follmann1991identifiability} extended this work to mixtures of logistic regression models with random intercepts. \citet{butler1997consistency} considered a latent linear model for binary data with any class of mixing distribution and provided sufficient conditions for identifiability of the fixed effects and mixing distribution. The main result of this work for our current discussion is easily summarized: not all mixture models are identifiable. This implies that the parameters can not be uniquely estimated for all mixture models . Of course, it is not very hard to see why this is the case: suppose we consider a simple mixture of two Bernoulli random variables:  $p(y = 1) = \alpha p_1^y (1-p_1)^{(1-y)} + (1-\alpha) p_2^y (1-p_2)^{(1-y)}$ where $0 < \alpha < 1$ is specifies the mixture probability and $p_1$ and $p_2$ quantify the probably of success $y=1$ for each of the two Bernoulli processes. Clearly, the parameters $\alpha, p_1$, and $p_2$ are unidentifiable: informally, we could say that multiple combinations of parameters would each maximize the likelihood of a given dataset. To illustrate, consider the following five observations: $1,0,0,0,1,1$. Any choice of parameters for which $\alpha p_1 + (1-\alpha) p_2 = .5$ holds maximizes the likelihood and thus there is an infinite number of parameter sets that yield the same distribution.

The example above immediately illustrates that the mixture of logistic regressions is not always identifiable: if we consider a finite mixture of two logistic regression models with Bernoulli observations where each model contains only one intercept, the situation is exactly the same as in our example. However, a number of mixtures of logistic regression models have been are identifialbe, and sufficient conditions for identification have been studied \citep{follmann1991identifiability}. Identifiability can be obtained in two ways for the finite mixture of logistic regression model. The first is by considering Binomial data generation instead of Bernoulli: if we manage to meaningfully group our observations, this can ensure identification. If we happened to know that  the five observations originate from two Binomial processes such that $1_a, 0_a, 0_a, 0_b, 1_b, 1_b$ where the subscript identifies distinct groups, than we could obtain the maximum likelihood estimate of $p_1 =1/3$, $p_2=2/3$, and $\alpha = .5$. More generally, having $m$ observations for each Binomial process is sufficient to identify $k \leq \frac{1}{2}(m+1)$ mixture components. The presence of such batched or grouped data also ensured identifiability in our coin tossing example used to illustrate the general EM algorithm.

Unfortunately, this approach is of little use in the streaming data scenario we have been focussing on; it would entail storing the groupings in a way which would complicate the fitting algorithm. Luckily, however, having repeated observations is not the only sufficient condition for a mixture model to be identifiable (and note that it is not a necessary condition). A mixture of logistic regression models can also be identified, informally, when the ÒpatternÓ within each individual logistic regression is clear enough. If enough unique values are available for the independent variables, then this can be used as a sufficient criteria for identifiability. \citep{follmann1991identifiability} showed that for a single independent variable having $q$ unique values, a mixture with $k \leq \sqrt{q+2}-1$ can be identified, at least if each of the logistic regressions is itself also identified (in the same way as a standard non-mixture logistic regression cannot be fit without additional constraints  if the number of parameters is larger than the number of observations). Hence, when treating the observations as Bernoulli, sufficient criteria for identifying mixtures logistic regression models can also be obtained in the data stream. However, we should note that the number of components that can be identified as more unique values of an independent variable are observed grows slowly: for $100$ unique values, \citet[][]{follmann1991identifiability} Theorem only guarantees sufficiency for identifying $9$ mixture components. Since we are considering large, continuous, data streams here, we consider the situation of more than $100$ unique values of an independent variable quite likely, but we must still caution for over-enthusiasm when choosing a high number of mixture components. A formal discussion of identifiability for mixture models in general can be found in \citet{titterington1985statistical} or \citet{fruhwirth2006finite}, while \citep{follmann1991identifiability} discuss specifically the mixture of logistic models.

\subsection{Convergence}

Even when a model is identified, one could wonder whether the procedure used to estimate the model parameters (whether iterative or sequential) converges: e.g., whether the parameter values can theoretically be guaranteed to end up in a position where the likelihood function is maximized (or at least the procedure finds a local maximum). The suggested procedure for oFMLR is challenging in this case since it combines stochastic gradient descent, with an online version of the EM algorithm, both of which might or might not converge. However, convergence of both the EM algorithm (and online versions there-of) and SGD have been widely studied. 

For SGD the study of convergence, when assuming a stationary (e.g., non-time changing) process revolves around finding a learn rate $\gamma$ that is both large enough for the parameter estimates to change sufficiently in the face of new evidence, while also small enough for the parameters to stabilize. This intuition can be formalized using the following criteria:
$\sum \gamma_t = \infty$ and $\sum \gamma_t^2 < \infty$. These objectives can be attained by using a gradually decreasing learn
rate. Learn rates such as $\gamma_t = \gamma_0t^{-\alpha}$, $\alpha \in (1/2, 1]$ have frequently been used \citep{kubrusly1973stochastic, cappe2009line}. However, such a decreasing learn rate is not the only feasible choice; constant (but small) learn rates effectively give more weight to recent observations than older ones, and thus fixed learn rates can be considered a sort of smooth moving window. Especially when the data generating process is expected to change over time, such a fixed learn rate might be preferred over learn rates that are guaranteed to converge but assume stationarity. This means that a small fixed learn rate combines online estimation with a smooth moving window approach.

For oFMLR, if the individual SGDs converge, we must ask whether the latent variable scheme used in the EM algorithm converges. This has also been studied for a large number of cases, specifically by \citet{cappe2009line} who considered the convergence of computationally feasible EM approximations similar to the online version suggested here.We omit the technical details but caution the reader to assess convergence by at least studying the traces of the fitted model parameters; an approach for oFMLR will be illustrated below.

In practice, offline model convergence is often determined by looking at  the changes of model parameters over different iterations. The notion of multiple iterations is absent in the online or streaming context, which means that traditional convergence statistics can not be used. However, we propose a similar approach, looking at the following statistic (included by default in our [R] package) when assessing convergence of oFMLR in a data stream:
\begin{eqnarray}
\bar{\delta}_{\lVert \vect{\theta} \rVert} 	&:=& \frac{ | \lVert \hat{\vect{\theta}} \rVert_t - \lVert \hat{\vect{\theta}} \rVert_{t-1} |  - \bar{\delta}_{\lVert \vect{\theta} \rVert}  }{ \texttt{min}(t, m)}.
\end{eqnarray}
Here we use $\lVert \hat{\vect{\theta}} \rVert_t$ to denote the $l_2$ norm (or Euclidian distance / length, $\lVert \vect{a} \rVert = \sqrt{a_1^2 + a_2^2 + \dots}$) of the parameter vector $\hat{\vect{\theta}}$ at a specific data point $t$, and $\bar{\delta}_{\lVert \vect{\theta} \rVert}$ to denote the moving average of the difference of this length compared to the length of the parameter vector at the previous datapoint $t-1$. The term $\texttt{min}(t, m)$ in the denominator ensures that we are looking not just at the average change in $l_2$ norm during the full data stream, but rather at a moving window; once the number of observations $t$ exceeds the window length $m$, older datapoints are smoothly discarded. This is similar to the effect of choosing a small, but fixed, learn rate, as in the SGD example discussed earlier. We will examine the performance of this convergence metric below.

\subsection{Interpretation of the Model Parameters and Choosing $K$}
\label{sec:choosek}

One of the advantages of oFMLR is that it provides a meaningful interpretation of the parameters within each cluster, which is useful for policy making. Hence, contrary to more Òblack boxÓ approaches to clustering \citep[Cf.][]{Chipman2007, Chipman2010}, the current approach allows researchers to not only segment customers, but also to interpret these segments. The first meaningful interpretation can be derived from the estimates of vector $\vect{\alpha}$. Each element $k=1, \dots, k=K$ of $\alpha$ represents the estimated proportion of customers belonging to this specific segment. Inspecting the elements of $\vect{\alpha}$ can give one direct information regarding customer segments and their size. In addition, estimating these segments in a data stream provides researchers with the opportunity to inspect the dynamics in customer segments: if the values of the elements of $\vect{\alpha}$ change over time, then the behavior of different segments is apparently not stable.

Researchers can also directly interpret the estimated regression coefficients $\vect{\beta}_k$ within each cluster. These estimates are standard logistic regression estimates, and can thus be interpreted as such. Within a specific cluster, the vector of estimated coefficients, $\beta_{0k}, \beta_{1k}, \dots, \beta_{pk}$, where $p$ is the number of predictors in the model, is available during in the data stream. These coefficients can be interpreted in terms of familiar log-odds, and one can directly inspect which variables play the largest role in the behavior within a segment.

In this discussion, we have not considered the choice of  $K$: it was treated as given. However, in reality, the number of clusters $K$ is unknown before the start of the analysis. The choice of $K$ might be strictly guided by theory, which allows the analyst to choose $K$ in advance. However, in conventional, offline, analyses  $K$  is often determined ad hoc by fitting models with different choices of $K$ and inspecting their model fit and estimated parameters. The latter is quite easily incorporated when using oFMLR: one can simply estimate, in parallel, models with different choices of  $K$ (see simulation results below).

Formally choosing the ``best'' number of components $K$ is more challenging: although we can easily fit multiple models with different choices of $K$ in parallel, formal tests to determine the value of $K$ in the face of streaming data are still actively being developed \citep{zhou2005streaming}. The traditional, offline,  methods for selecting the number of components mostly rely on comparing either the likelihood or the $AIC$ or $BIC$ values of the competing models \citep{titterington1985statistical, fruhwirth2006finite}. The likelihood $L(\hat{\vect{\theta}} | \mathcal{D})$ (or log-likelihood $l(\hat{\vect{\theta}} | \mathcal{D})$ of the mixture of logistic regression is easily computed for a static dataset (see also Eq. \ref{likelihood:ofmlr}) and a set of parameter estimates. Subsequently computing the $AIC$ or $BIC$ to enable model comparisons is straightforward:
\begin{eqnarray}
AIC &=& 2k - 2 \ln(\hat{L}) \\  
BIC &=& -2 \ln(\hat{L}) k \ln(n)
\end{eqnarray}
where $n$ is the total number of observations, $k$ is the number of parameters in the model, and for conciseness $\hat{l}$ is chosen to denote the maximized value of the likelihood function.

The corresponding quantities are however not easily defined in a continuous data stream. Computing the full likelihood $\hat{L}$ is not feasible if the data set is large, since this requires re-evaluating the log-likelihood of the full data set for the current state of the parameters $\hat{\vect{\theta}}_t$. Going back through all the data in retrospectively would defeat the purpose of fitting the model in the stream. For oFLMR we therefore chose a slightly different approach, one that is---to the best of our knowledge---novel. We compute the streaming average log-likelihood of each datapoint given the state of the parameters at that moment for a smooth window of pre-specified size $m$:
\begin{eqnarray}
\bar{l} & := &  \frac{ \hat{l}_t - \bar{l} }{ \texttt{min}(t, m)},
\label{eq:onlineloglik}
\end{eqnarray}
\noindent
where $\hat{l}_t$ denote the log-likelihood of the datapoint arriving at time $t$. Thus, $\bar{l}$ quantifies the average log-likelihood of each of the streaming datapoints. However, this notation hides the fact that the value of $\hat{\vect{\theta}}_t$ used to compute $\hat{l}_t$ changes at each timepoint and thus provides a stochastic approximation. The streaming average log-likelihood $\bar{l}$ provides a quantification of the fit of the model for the latest $m$ points in the data stream; we discuss the behavior of this metric below. From this metric, we can derive streaming approximations to the $AIC$ and $BIC$, respectively:
\begin{eqnarray}
sAIC &=& 2k - 2 \bar{l} \   \texttt{min}(t, m) \\   
sBIC &=& -2 \bar{l} \  \texttt{min}(t, m) \  p \ln(m)
\end{eqnarray}
which can be used to compare models with different numbers of parameters fitted in parallel during the  data stream.

\subsection{Alternative (Bayesian) Approaches}
\label{sec:Bayes}

In our presentation of oFLMR we have focussed on a frequentist approach. However, ever since the introduction of finite mixture modeling, authors have considered Bayesian approaches for estimating parameters \citep[see][for an introduction]{titterington1985statistical}. As is often the case, the Bayesian approach for parameter estimation is simple:
\begin{eqnarray}
p(\vect{\theta} | \mathcal{D}) 	&=& \frac{P(\vect{\theta} | \mathcal{D}) p(\vect{\theta})}{p(\mathcal{D})} \\
						&\propto& L(\vect{\theta} | \mathcal{D}) p(\vect{\theta})
\label{Eq:Bayes}
\end{eqnarray}
which indicates that we can obtain the joint posterior distribution of $\vect{\theta}$ by simply specifying a prior distribution $p(\vect{\theta})$, after which, if we are interested, we can obtain point estimates of $\hat{\vect{\theta}}$ by summarizing the posterior $p(\vect{\theta} | \mathcal{D})$ in some way, often taking its expectation or its maximum value \citep{Gelman2008}. This simple formula seems to indicate that we only need to specify a prior distribution to compute $p(\theta | \mathcal{D})$ since we have already worked out the likelihood. Furthermore, one should note that the Bayesian framework potentially also offers tremendous opportunities for dealing with data streams \citep{Opper1998a} since the following holds,
\begin{eqnarray}
p(\vect{\theta} | \mathcal{D}_{t+1}) 	&\propto& L(\vect{\theta}_{t+1} | \mathcal{D}_{t+1}) p(\vect{\theta} | \mathcal{D}_t),
\label{Eq:BayesOnline}
\end{eqnarray}
\noindent
indicating that we can use the posterior at time $t$ as the prior for time $t+1$ and thus naturally update our beliefs as the data come streaming in. Using Eq \ref{Eq:BayesOnline} Bayesian parameter estimation is naturally carried out online!

Nevertheless, it is not as straightforward as it would appear. It has long been known that simple analytic treatment of even \ref{Eq:Bayes} is not possible when $L(\vect{\theta} | \mathcal{D})$ derives from a mixture model  \citep[see][and references therein]{titterington1985statistical}. It follows that implementing a Bayesian estimation procedure for finite mixture models is not straightforward when analyzing static data, let alone in the streaming data case we are considering. 

In recent years, Bayesian approaches that have been successful have relied on sampling methods and / or other types of approximations. MCMC sampling is popular \citep[see, e.g.,][]{Ryu2011}, and effective Markov Chain Monte Carlo (MCMC) samplers for finite mixture models have been presented  \citep{dippold2013model}. For the standard (offline) fitting of a Bayesian mixture model, software packages such as JAGS or Stan can be used relatively easily \citep{kruschke2014doing}. Approximations using variational methods have recently gained popularity, and offline versions can of this algorithm can easily be found  \citep{attias1999inferring}. 

However, only recently have authors started to consider online (or streaming) versions of these estimation methods: recent work on sequential MCMC \citep{kantas2009overview, scott2016bayes} and stochastic variational Bayes (VB) methods \citep{hoffman2013stochastic, tank2015streaming} suggests computationally attractive approximations for fitting complex Bayesian models in data streams. To the best of our knowledge, there are however no ready to use implementations of either sequential MCMC or stochastic VB approaches for the finite mixture of logistic regression models presented here; developing such implementations would be a great benefit.

\section{Evaluations of oFMLR}
\label{sec:empirical}

This section examines how oFMLR performs, using a simulation study and an empirical example. In each case we describe the data that are used, the resulting estimates, and we reflect on convergence and model selection. All the models presented here have been fit using the [R] oFMLR package that accompanies this article. The package allows easy fitting of mixtures of logistic regression models in data streams and allows one to compare multiple models in parallel that are fit to the same data stream.\footnote{is presumably not the programming language of choice for dealing with large data streams in practice. However, several tools that do handle large data streams, such as Apache Spark \citep[for example see][]{salloum2016big}, directly interface with [R].}

Before presenting the simulation study and our empirical example, we first describe the oFMLR package in a bit more detail and present some code examples to enable readers to directly use the package themselves. The following [R] code downloads and installs the package:
\begin{lstlisting}
> library(devtools)
> install_github("MKaptein/ofmlr")
> library(ofmlr)
\end{lstlisting}
Here, the first line is used to load the \texttt{devtools} package which allows---among many other utilities---[R] packages to be easily installed from \texttt{github}. The second line downloads and installs the oFMLR package (this line only needs to be run once) and the final line is used to load the package. After running these three lines of code, the package is ready for use and, as is standard in [R], one can start by browsing the documentation using \texttt{?ofmlr}.

Next, we present an example for fitting a simple model. The following code allows one to generate a mixture dataset and subsequently fit the oFMLR model to a simulated data-stream (note that the lines following the \# sign are comments):
\begin{lstlisting}
set.seed(123456)

# Use utility functions in package to generate data
n <- 10^5
data <- generate_mixture(n, 2, 2, beta=matrix(c(3,-2.5,-2, 5), nrow=2), ak=c(.3,.7))

# Create a new oFMLR object 
M <- online_log_mixture(2,2, trace=1000, ll.window=1000)

# "Stream" the data using a for loop
for(i in 1:n){
	M <- add_observation(M, data$data[i,1], data$data[i, -1], lambda=0)
}
\end{lstlisting}

The first line of the code sets the seed for reproducibility of the results. Next, we use the the \texttt{generate\_mixture} utility function included in the package to create a dataset consisting of $N=10^5$ observations with the number of mixture components $k=2$, the number of parameters per logistic regression model $p=2$, and with model parameters $\vect{\alpha} = (.3,.7)$, $\vect{\beta}_1 = (3, -2.5)$ and $\vect{\beta}_2 = (-2, 5)$. The next line is used to instantiate a new oFMLR object with similar properties (e.g., $k=2$ and $p=2$) and random starting values for $\vect{\alpha}$ and $\vect{\beta}$ where the latter are drawn uniformly between $-1$ and $1$; for the current choice of seed these starting values are $\hat{\vect{\alpha}} = (.31,.69)$, $\hat{\vect{\beta}}_1 = (-.65, .81)$ and $\hat{\vect{\beta}}_2 = (-.09, .74)$. The \texttt{trace=1000} argument in the call to instantiate the new oFMLR object ensures that a snapshot of the parameters ($\hat{\vect{\theta}}_t$) is stored every $1000$ datapoints as well as the values of $\bar{\delta}_{\lVert \vect{\theta} \rVert}$, $\bar{l}$, $sAIC$, and $sBIC$ respectively. Finally, the \texttt{lambda=0} argument sets the learn rate to the default decreasing learn rate $\lambda = n^{-\frac{1}{2}}$.

\begin{figure}
  \centering
    \includegraphics[width=0.7\textwidth]{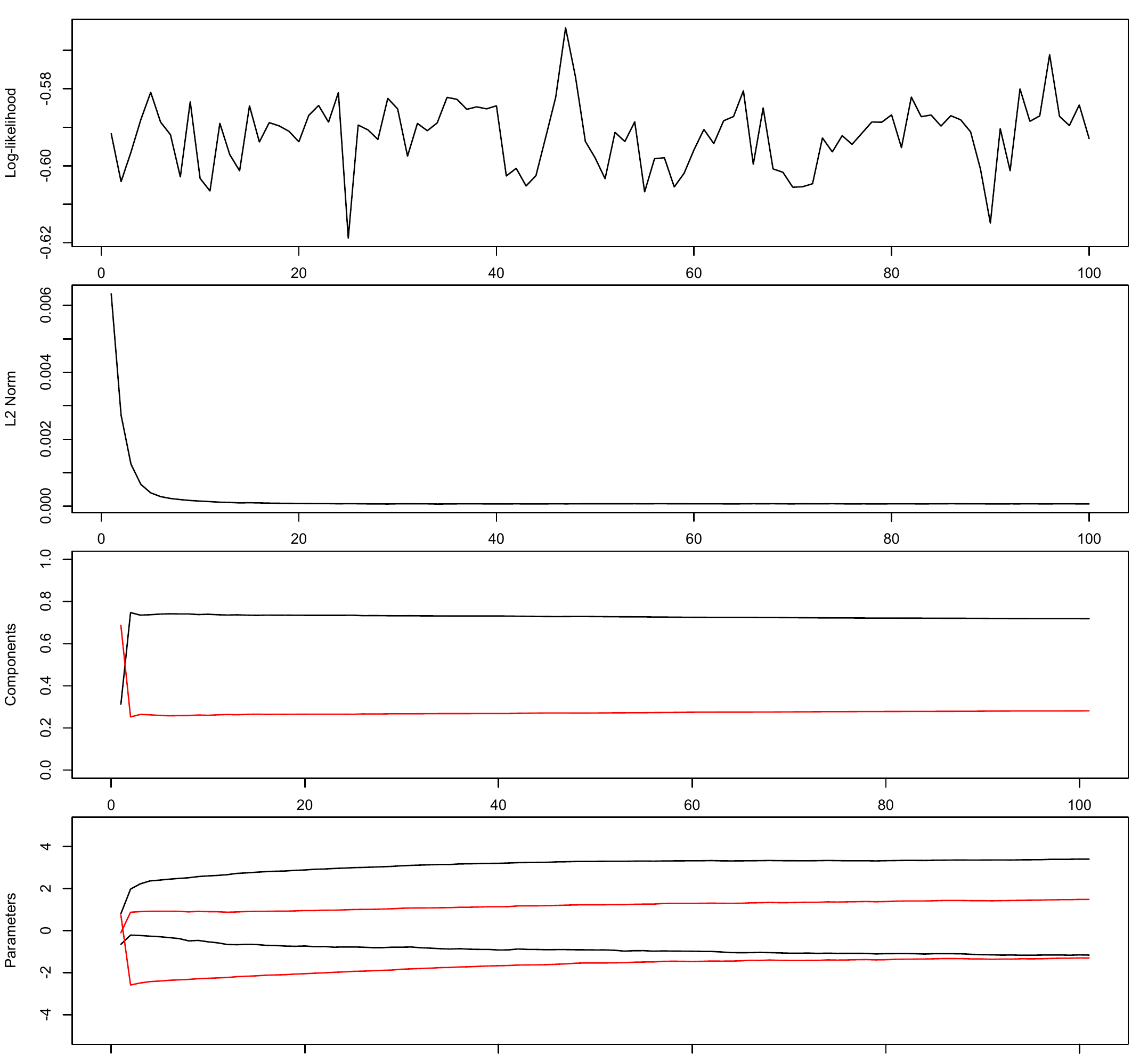}
      \caption{Trace of values of $\bar{l}$, $\bar{\delta}_{\lVert \vect{\theta} \rVert}$, $\hat{\vect{\alpha}}$, and $\hat{\vect{\beta}}$.}
      \label{fig:trace}
\end{figure}

After instantiating the model the \texttt{for} loop simulates a sequential run through the dataset. At each row of the dataset that is visited, the parameter values of the oFMLR model are updated. After this process has finished, a call to \texttt{summary(M)} produces the following output:
\begin{lstlisting}
Online fit of logistic mixture model (oFMLR)
Number of mixture components:  2 
Number of predictors (including intercept):  2 
Estimated cluster membership probabilities: 
[1] 0.719 0.281
Estimated coefficients:
      [,1] [,2]
[1,] -1.16  3.4
[2,]  1.49 -1.3
Total number of observations in data stream:  1e+05 
The current (streaming average) log likelihood is  -0.593 (-0.076), sAIC= 1197.197, sBIC= 1226.644 .
NOTE: The l2 norm of the parameters has changed by < 0.001 in the last update.
\end{lstlisting}
This shows that the parameter estimates are reasonable---although the values for $\hat{\vect{\beta}}$ still seem influenced by the starting values. The values of $\hat{\vect{\alpha}}$ however are very close to the true values.
We investigate the precision of the parameter estimates more thoroughly in the simulation study presented below. The ``NOTE'' at the end highlights that the parameter values have not changed much over the last data points. It is difficult to interpret the log-likelihood and the $sIAC$ and $sBIC$ in isolation; below we illustrate how these can be used for model comparisons.
Finally, the value between brackets after the $\bar{l}$ denotes the maximum average log-likelihood of the datapoints when using fixed cluster assignments based on the posterior probabilities $z_{it}$; this can be used for additional diagnostic purposes. Since we have traced the progress of the model fitting in the data stream, a call to \texttt{plot(M1, params=TRUE)} produces Figure \ref{fig:trace}, which gives an overview of the learn rate and convergence diagnostics.

In addition to fitting an individual model, the oFMLR package also allows comparisons of multiple models fit to the same datastream. The following [R] code instantiates multiple models and adds them to a model comparison object (which we will call a \texttt{multi\_online\_log\_mixture}) which is subsequently updated using a data stream:
\begin{lstlisting}
compare_models <- multi_online_log_mixture(online_log_mixture(2,1))
compare_models <- add_model(compare_models, online_log_mixture(2,2))
compare_models <- add_model(compare_models, online_log_mixture(2,3))
for(i in 1:n){
	compare_models <- add_observation(compare_models, data$data[i,1], data$data[i, -1], 0)
}
\end{lstlisting}
After running the above code, a call to \texttt{summary(compare\_models)} produces Table \ref{tab:compare}. Here the use of the sIAC and sBIC is clear: the model with two components, which fits the true data generating model, is preferred.\footnote{In practice we recommend running multiple models in parallel, each with different starting values, to asses model fit.} Having introduced the oFMLR package, we will investigate the parameter convergence and its practical applications   in the next section.


\begin{table}[ht]
\centering
\begin{tabular}{rlrrrrrrrr}
  \hline
 & M & k & p & ll & maxll & sAIC & sBIC & dNorm & n \\ 
  \hline
1 & M1 & 1.0 &  2 & -0.618 & -0.618 & 12359 & 12381 & 0.0034 & 100000 \\ 
  2 & M2 & 2.0 &  2 & -0.592 & -0.182 & 11853 & 11896 & 0.0001 & 100000 \\ 
  3 & M3 & 3.0 &  2 & -0.592 & -0.126 & 11863 & 11928 & 0.0001 & 100000 \\ 
   \hline
\end{tabular}
\caption{Table of model comparisons created by calling \texttt{summary()} on a \texttt{multi\_online\_log\_mixture} object.}
\label{tab:compare}
\end{table}

\subsection{Simulation Study: Parameter estimates}

This section examines the convergence of the oFMLR over a large series of models and for different sizes of data stream. We  simulate the previously mentioned scenario in which the observed clicks on an advertisement that originate from two homogeneous clusters of customers: one sensitive, the other seeking exclusivity. Once again we set $\vect{\alpha} = (.3,.7)$, $\vect{\beta}_1 = (3, -2.5)$ and $\vect{\beta}_2 = (-2, 5)$ as the true model parameters and simulate $n=10^5$ observations from this model. We than draw the values of the dependent variable (the $x_t$'s) randomly between $-5$ and $5$ and subsequently simulate an observation from the model. Thus, to simulate an observation we first draw a cluster membership ($z_t \sim \mbox{Bern}(.2)$, and then fill in a value for $x_t \sim \mbox{Unif}(-5,5)$ in the respective data generating model. We repeat this process for $m=100$ simulations. 

Next, we fit both an offline version of a $K \in \{1,2,3\}$ cluster model using standard EM (as implemented in the \texttt{mixtools} package in [R] \citep{benaglia2009mixtools}) and an online version using the oFMLR package. For both the online as well as the offline version, we need to specify starting values for both $\vect{\alpha}$ and $\vect{\beta}_k$. We choose $1/K$ as the value for each element of $\alpha$, and a random draw $d \sim \mbox{Unif}(-1,1)$ for each of the elements of $\beta$. Furthermore, for the offline version, we choose a fixed number of $100$ iterations to allow the EM algorithm to converge.\footnote{A larger number of iterations might improve the parameter estimates in some runs. A smaller number obviously decreases running time. However, we feel that a fixed---albeit relatively large---number of iterations provides a more valid comparison.} Finally, for the online version, we choose a fixed learn rate: $\gamma = .1$.

gives an overview of the obtained parameter estimates for different lengths of the data stream. The mean and the standard error (SE) as computed over the $m=100$ simulations are presented.\footnote{Note that the order in which the clusters are identified is not the same in every simulation run. For interpretation purposes, we order both the elements of $\vect{\alpha}$ and the components of $\vect{\beta}$.} A number of things are quite striking: first, both of the methods largely fail on the smaller data streams (or at least have very high standard errors on the parameters). Both methods seem to work relatively well for medium size streams, and the resulting estimates are close to the true data generating model. Both methods perform well for large streams; the signs of all the coefficients are already properly estimated at $n=10^4$. The current simulations further show that for $K=3$, and hence a number of clusters that is too high compared to the data generating model, we find two clusters that have similar estimates for $\hat{\vect{\beta}}_k$ when using oFMLR. As mentioned earlier, this can be used to select the number of clusters when analyzing a data stream. This does not seem to work for the offline method, which is unable to converge---using the standard convergence criteria in the \texttt{mixtools} package---when the number of clusters is too high. Finally, for an online simulation with $n=10^6$  the estimated coefficients using oFMLR are $\alpha = (.32,.68)$, $\beta_1 = (2.8, -2.3)$ and $\beta_2 = (-1.9, 5.3)$ (all SEs $<.01$) (This was not presented in the table since the offline version takes too much time). These values are very close to the true data generating values. At small sample sizes it is clear that the starting values for the SGD still play a large role in the parameter estimation; however, for large data streams ($n>10^5$) our oFMLR seems both fast and accurate.

 \begin{table}
 \centering
 \begin{tabular}{llHrrr}
		\hline
        		&             & $10^2$                 & $10^3$               & $10^4$               & $10^5$               \\ 
		\hline
Offline        & $\beta_1$   	&         0.03 (0.02)               &      0.04 (0.01)                &      0.03 (0)                &        0.04 (0)              \\
        &             		&          0.29 (0.01)        	&        0.27 (0)         	&       0.27 (0)          &         0.27 (0)        \\
	\hline
        	& $ \alpha$   &      0.38 (0.01)              &       0.44 (0)               &       0.46 (0)               &        0.46 (0)              \\
	&    		&            0.62 (0.01)            &          0.56 (0)            &         0.54 (0)             &          0.54 (0)            \\
        	& $\beta_1$   &         27.47 (11.2)        &          -0.59 (0.22)      &       0.08 (0.06)               &       0.21 (0)               \\
        	&             &           -11.55 (5.11)            &       3.25 (0.74)          &        0.3 (0.19)        &          -0.16 (0)       \\
        	& $ \beta_2$  &       -30.42 (17.39)       &         -16.17 (6.33)    &        -1.44 (0.06)              &        -1.5 (0.03)              \\
        	&             &          113.61 (45.12)          &       52.16 (20.41)     &          6.07 (0.2)       &       6.41 (0.07)          \\
	\hline
       	 & $\alpha   $ &        0.2 (0.01)        	&       0.23 (0)               &       0.22 (0)               &        0.23 (0)              \\
	& 		  &         0.26 (0.01)          	&       0.28 (0)               &       0.28 (0)               &           0.26 (0)           \\
	&    		&          0.53 (0.01)           	&        0.49 (0.01)             &     0.5 (0)                 &         0.52 (0)             \\
        	& $ \beta_1 $ &       37.82 (26.79)   	 &          0.27 (0.07)            &        0.31 (0.04)              &       0.07 (0.03)               \\
       	&             &            -11.21 (11.12)       	&    0.44 (0.11)             &      0.37 (0.07)           &     0.03 (0.03)            \\
        	& $\beta_2 $  &     53.64 (35.03)     	&         0.27 (0.07)             &     0.29 (0.02)                &      0.34 (0.04)                \\
        	&             &            -31.65 (12.5)       	&         0.2 (0.39)        &        -0.34 (0.01)         &          -0.28 (0.04)       \\
        	& $ \beta_3 $ &     -34.72 (27.09)           	&      -16.8 (6.07)                &       -2.19 (0.08)               &       -1.69 (0.06)               \\
        	&             &            184.71 (51.74)       	&      52.02 (16.34)           &    7.97 (0.17)             &      7.14 (0.13)           \\
	\hline
Online        	& $\beta_1$   &        0.03 (0.03) &         0.05 (0.03)             &      0.01 (0.02)                &        0.02 (0.03)              \\
        	&             	  &        0.32 (0.02)   &          0.31 (0.02)       &     0.33 (0.02) &      0.32 (0.03)       \\
        \hline
        & $ \alpha$   	&          0.48 (0)              &       0.44 (0)               &        0.4 (0)              &        0.35 (0)              \\
        & 		   	&          0.52 (0)              &        0.56 (0)              &          0.6 (0)            &        0.65 (0)              \\
        & $\beta_1$   	&          0.01 (0.02)              &       0.12 (0.02)               &     0.42 (0.02)                 &       0.99 (0.04)               \\
        &             		&           0.19 (0.02)        	&        -0.27 (0.02)         &        -0.55 (0.02)         &      -0.97 (0.03)           \\
        & $ \beta_2$  	&          0.01 (0.02)         	&       0.02 (0.02)          &         -0.76 (0.02)        &     -1.71 (0.03)            \\
        &             		&           0.41 (0.02)             &         1.7 (0.02)             &         3.77 (0.02)             &          5.6 (0.02)            \\
        \hline
        & $\alpha   $ &       0.31 (0)           &       0.31 (0)               &         0.3 (0)             &         0.3 (0)             \\
        & 		&           0.33 (0)           &          0.33 (0)            &         0.33 (0)             &           0.33 (0)           \\
        &		 &           0.35 (0)           &          0.36 (0)            &        0.37 (0)              &         0.37 (0)             \\
        & $ \beta_1 $ &        0.01 (0.01)     &       0.07 (0.02)               &         0.63 (0.03)             &      2.28 (0.11)                \\
        &             &        0.21 (0.02)           &        -0.44 (0.04)         &      -0.83 (0.09)           &      -1.8 (0.17)           \\
        & $\beta_2 $  &       0.01 (0.01)    &        0.03 (0.01)              &       -0.37 (0.04)               &        -1.46 (0.11)              \\
        &             &          0.3 (0.01)         &       0.98 (0.04)          &        2.53 (0.11)         &      4.28 (0.18)           \\
        & $ \beta_3 $ &        0.02 (0.01)   &         0.01 (0.01)             &       -0.63 (0.02)               &         -2.01 (0.02)             \\
        &             &        0.36 (0.01)         &      1.29 (0.03)           &       3.11 (0.02)          &      5.15 (0.02)           \\
        \hline
     \end{tabular}
    \caption{Results of simulation Study 1. Data generated using a true model with parameters $\alpha = (.3,.7)$, $\beta_1 = (3, -2.5)$ and $\beta_2 = (-2, 5)$.}
    \label{tab:sim1results}
    \end{table}

\subsection{Empirical Evaluation: Practical utility}
\label{sec:empirical}

To demonstrate the utility of our proposed method for marketing applications we use the oFMLR [R] package to do a \emph{post-hoc} analysis of an existing large dataset of online consumer browsing behavior. The dataset consists of $n=105502$ observations of the web-browsing behavior of prospective customers of a large insurance company (the data was collected between March $1$ and June $30$, 2016). The dataset supplied by the firm, which wishes to remain unidentified, contains records of unique customers, identifying the number of pages they visited in their first visit to the  the insurance companyÕs website, the total time spent browsing in this session, and whether or not the customer ended up purchasing a product. These behavioral measures were later augmented by the company with an orientation ÒscoreÓ. Based on the content of the pages that are visited, the insurance company enriches their database by dividing customers into two categories: those who are merely \emph{looking} for information regarding the company and different  insurance types, and those who seem to be interested in \emph{purchasing} one of their products, determined by the fact that they visit one or more of the actual product information pages. Table \ref{tab:overview} presents some descriptives to further illustrate the current dataset.

\begin{table}[htpb]
\centering
\begin{tabular}{llrrrr}
\hline
         Variable & Description & Min & Max & Mean & Median \\
         \hline 
Purchase & Did the customer buy? &  0  & 1    &  0.037    &  0      \\
Time     & Total session time &  $<$1   & 1190    &  283    &  180      \\
Pages    &  Number of pages visited in session & 1  & 19    &   4.4   &   4     \\
Orient   & ``Looker'' or ``purchaser'' &   0  &  1   &  .53    &   1 \\
\hline   
\end{tabular}
\caption{Descriptives for the consumer browsing behavior dataset. Note the fairly low purchase (or convergence) rates as common in many online marketing applications. The total number of observations was $n=105502$ and the data was collected between March and June 2016.}
\label{tab:overview}
\end{table}

Interestingly, a separate analysis, using a simple logistic regression model with ``time'' and ``pages'' as predictors, shows that the effect of the time that consumers spend on the page is positive for those that are merely ``looking'', while it is negative---albeit not significantly---for those customers that seem to be interested in purchasing. Table \ref{tab:look} presents the results of these two independent analyses. The results are fairly intuitive: when visitors that wish to purchase a product spend more time, they are most likely experiencing barriers in the check-out process, while for visitors who are merely ``lookers'', a longer session probably relates to an increased interest.

\begin{table}[ht]
\centering
\begin{tabular}{lrrrrr}
  \hline
 & & Estimate & Std. Error & z value & Pr($>$$|$z$|$) \\ 
  \hline
 \textbf{Lookers}	& (Intercept) & -3.2445 & 0.0261 & -124.42 & 0.0000 \\ 
  	& Time & 0.6572 & 0.0183 & 35.83 & 0.0000 \\ 
  	& Pages & 0.4527 & 0.0161 & 28.20 & 0.0000 \\ 
   \hline
 \textbf{Purchasers} & (Intercept) & -5.6142 & 0.0726 & -77.30 & 0.0000 \\ 
  	& Time & -0.0425 & 0.0898 & -0.47 & 0.6356 \\ 
  	& Pages & 0.0135 & 0.0943 & 0.14 & 0.8858 \\
\hline
\end{tabular}
\caption{Comparison of simple logistic regressions to examine the effect of timing and the number of visiting pages on those that are (ad-hoc) qualified as either ``Lookers'' or ``Purchasers'' based on the content of the visited webpages.}
 \label{tab:look}
\end{table}

To evaluate the use oFMLR, we fit multiple logistic mixture models to the data while simulating a data-stream; hence, we effectively ``replay'' all the observations in chronological order and fit a number of competing models with different starting values and different number of components to the generated stream. Furthermore, we assume that the orientation score is unknown at the time the data arrives; these scores are determined ad hoc by the insurance company after a lengthy analysis of the urls which the consumers visited. The code below details how the ofmlr package was used to simulate the data stream and fit multiple models:
\begin{lstlisting}
# open the dataset
data <- read.csv("session_data.csv")

# use the ofmlr package and add multiple models
library(ofmlr)
models <- multi_online_log_mixture(online_log_mixture(3,1))
models <- add_model(models, online_log_mixture(3,1))
models <- add_model(models, online_log_mixture(3,2))
models <- add_model(models, online_log_mixture(3,2))
models <- add_model(models, online_log_mixture(3,3))
models <- add_model(models, online_log_mixture(3,3))

# Simulate the data stream
for(i in 1:nrow(data)){
	models <- add_observation(models, data[i,1], c(1,data[i, 2:3]) )
}
\end{lstlisting}
This code fits six distinct models in the simulated stream: we fit models with different (random) starting values that contain $3$ predictors (the intercept, the effect of ``time'', and the effect of ``page''), and either $1$, $2$, or $3$ mixture components; given the large number of unique values for the predictors, we should have sufficient data to identify models of this size. The \texttt{for} loop effectively ``replays'' the incoming data in chronological order. We fit the model after standardizing (computing $z$-scores) for the variables ``time'' and ``pages'' to prevent numerical instability.

The output of the call to \texttt{summary(models)} is presented in Table \ref{tab:results}. It is clear that the moving-average log-likelihoods of the two models with the same number of components but different starting values do not differ from each other much and that the average change in the norm of  the parameter vector is small, indicating reasonable convergence for each of the six models. When comparing the models based on the $sAIC$ and $sBIC$, we see that the single component model is preferred; perhaps the mixture component is either a) too small to identify, or b) the estimated negative coefficient of time for the ``lookers'' is not sufficient to affect the estimates for the two component model (in fact, this estimated coefficient was not significantly different from $0$ in the separate analysis). However, when looking at the maximum log-likelihoodÑwhich presents the mean log-likelihood for each datapoint when each datapoint is attributed to its highest posterior component---one would be inclined to prefer models $3$ or $4$. Both of these two-component models have a high value on this metric. This provides some evidence (albeit weak) in favor of the two-component model. Inspecting the model parameters of model $3$ shows a large cluster (76\%) of customers for whom the effect of time is positive, and a smaller one (23\%) for whom the effect is instead negative; at least qualitatively, this replicates the findings for the separate analysis, which included  the omitted variable ``orientation''. Hence, though the signal in the current dataset is weak, fitting multiple logistic mixture models in parallel can shed light on possible differences between clusters of consumers. Again, it is intuitive to identify clusters of visitors based on the time spent during their visit to the website: time on the page can be a proxy for general interest and thus have a positive effect, but it could also highlight usability issues with the website, in which case it will likely have a negative effect for consumers  trying to make an actual purchase.


\begin{table}[ht]
\centering
\begin{tabular}{rlrrrrrrrr}
  \hline
 & M & k & p & ll & maxll & AIC & BIC & Norm & n \\ 
  \hline
  2 & M1 & 1 & 3 & -0.1003 & -0.1003 & 208.60 & 228.23 & 0.0031 & 105502 \\ 
  3 & M2 & 1 & 3 & -0.1003 & -0.1003 & 208.60 & 228.23 & 0.0031 & 105502 \\ 
  4 & M3 & 2 & 3 & -0.1079 & -0.0416 & 231.87 & 271.13 & 0.0015 & 105502 \\ 
  6 & M4 & 2 & 3 & -0.1060 & -0.0539 & 227.96 & 267.22 & 0.0018 & 105502 \\ 
  7 & M5 & 3 & 3 & -0.1071 & -0.0588 & 238.25 & 297.14 & 0.0011 & 105502 \\ 
  9 & M6 & 3 & 3 & -0.1045 & -0.0967 & 233.04 & 291.93 & 0.0020 & 105502 \\ 
   \hline
\end{tabular}
\caption{Comparison of \emph{oFMLR} models with 1-3 mixture components fit to the consumer browsing behavior data in simulated a data stream.}
\label{tab:results}
\end{table}

\section{Discussion}

This article introduced a method----and associated [R] package---for the analysis of customer responses arriving in a continuous data stream (in this case, clicking behavior) The method is an online or row-by-row implementation of the EM algorithm to fit a finite mixture of logistic regression models. These models have been widely investigated, and we implement a specific version of the more general online EM algorithm discussed by \citet{cappe2009line}. Our specific implementation of the finite mixture of logistic regression models, and the software provided, make the current work especially suited for analyzing consumer behavior arriving in large and continuous data streams. We have detailed the challenges facing us when analyzing customer data that arrive in high velocity data streams and we have explained how the use of online estimation methods can help marketing scholars deal with the large streams of data that originate from online click streams. We have also discussed the concepts behind both SGD and the EM algorithm. SGD is an optimization method with many applications for fitting models in data streams: its importance for future work in analyzing online advertising data can hardly be overstated. The EM algorithm is a general algorithm for the dealing with missing data problems. In many cases, the EM algorithm can be transformed into an online version, thus providing a wealth of methods for analyzing data streams. Finally, we have discussed both identification and convergence issues, as well as linking our work to Bayesian approaches that have a similar direction. We hope the current paper will be instructive for readers unfamiliar with the analysis of large continuous data streams, as well as introducing our oFMLR model. 

We envision using oFMLR (or other related statistical models that can be fit online) in order to continuously monitor the effects of online marketing campaigns. We can encode both features of the marketing messages (the product being displayed, the shape, the persuasive appeals being used, etc), as well as features of the customers. The data stream that results from the click-behavior of customers on the messages or surrounding products can be analyzed continuously using multiple versions of oFMLR running in parallel (with different choices of $K$ and different starting values). The estimated mixture probabilities will provide direct and real-time feedback to policy makers about homogeneous clusters within the target audience, and the estimated coefficients within each cluster can be used to interpret these findings qualitatively: as our empirical example illustrated, the analysis can highlight both positive and negative effects of a single predictor. Interpreting such differences between clusters can inspire different courses of action: for example, in our empirical example one imagines a redesign of the website could improve its usability and make purchasing simpler.

Both the logistic regression model and its finite mixture extension (and variants of it) have been used for numerous applications in marketing research \citep[e.g.,][]{zhang2004customizing, van2007new, schmittlein1994customer, west1997comparative}. These models have been applied for a wide range of purposes:  to understand and cluster customersÕ online behavior,  to understand new product diffusion, and to model consumer choice. As such, the models are flexible and have a tremendous potential for applications in marketing; even recent developments in machine learning still heavily rely on the logistic regression model \citep[see, e.g.,][]{he2014practical}. Technological advances have created new measurement opportunities in all areas of marketing, and marketing researchers are increasingly confronted with high volume or high velocity streams. We hope the current work will contribute to the use of logistic regression models  to understand such continuously collected data. We believe that analysis practice will change in the coming years: when confronted with continuous data streams, a continuous analysis---such as demonstrated in our empirical example---can inspire continuous new marketing policies. Using fixed learn rates---and thus effectively ``forgetting'' older data---we can approach the estimates based on multiple parallel statistical models, providing a continuous source of information for policy decisions rather than testing specific hypotheses at specific points in time. The data collection is never finished, nor are our attempts to optimize our marketing  policies.

Admittedly, this paper has only discussed the Bayesian paradigm very briefly. This is by no means intended to discourage a Bayesian treatment of modeling continuous data streams; for example, the relatively recent Probit regression approach presented by \citet{graepel2010web} provides an extremely usable, scalable, and fully online approach to modeling binary outcomes using a single component. We would welcome the development of similar online approaches for mixture models. Large steps are also being made in statistics and computer scienceÑas witnessed by the recent work on stochastic variational inference and sequential MCMC \citep[e.g.,][]{tank2015streaming,scott2016bayes}.  In the years to come, these approaches surpass the frequentist treatment presented here because of the inherent benefits of the Bayesian approach when quantifying uncertainty in the estimated parameters. However, we have focussed on what we believe is still the most common approach in marketing practice; we hope the current work contributes to further developments in both Frequentist and Bayesian methods to deal with high velocity continuous data streams.

This paper has solely focussed on learning from a data stream case-by-case. However, in practice, models that are fit online might benefit from a hybrid approach in which either batches of datapoints are used to compute updates of the parameters or in which an offline analysis of a static dataset is used to determine the starting values for a model that is subsequently updated in a data stream.  Both of these approaches warrant further investigation. Such hybrid approaches have been specifically explored for the EM-algorithm (see \citet{neal1998view} and \citet{liang2009online} for examples). Finally, this paper introduced several diagnostic tools geared specifically towards the analysis of a data steamÑfor example, the online moving average of the log-likelihood as presented in Eq. \ref{eq:onlineloglik}; these proposals need to be studied in more detail.

In this paper, we demonstrated the performance of oFMLR in a simulation study, as well as applying it to an empirical dataset. We also provided an easy to use [R] package to fit the oFMLR model. While the results of our simulations are promising, and the reduction of computation time that is obtained by using online estimation is very appealing, we do have to stress that finite mixtures  of logistic regression models do not always converge properly: this is true both for the online   and the offline versions. Furthermore, especially in the online case, the analysis is influenced by the starting values, and, without substantial domain knowledge, it is hard to establish identifiability criteria at the start of the data stream. Hence, the analyst should---as always---be careful interpreting the results, and should seek additional methods to validate the clustering that is found when analyzing a data stream using oFMLR. In practice, we recommend fitting multiple models (for different choices of $K$) online, and using an online bootstrap method, such as suggested by \citep{Owen2012}, to quantify the uncertainty in the estimated model parameters. Combined with a long data stream, $T > 10^5$, this will ensure that our proposed method can be used efficiently and responsibly to identify clusters of customers based on click-stream data.

We believe that oFMLR provides an addition to the toolbox of statistical methods that is available to marketing scholars and practitioners when analyzing data streams.We also hope to have  provided additional understanding regarding a) the basic conceptual methods of dealing with high volume and high velocity data streams, and b) the methodological building blocks (SGD and EM) that were used to develop oFMLR.

\section*{Acknowledgements}%
M.K. would like to gratefully acknowledge the support of Dr. Joris Mulder, Prof. Dr. Dean Eckles, and Prof. Dr. Petri Parvinen for their insightful discussions during the preparation of this manuscript. Furthermore, M.K. would like to thank Hans van der Poel and Dave Kruizinga for supplying the data for the case study is Section \ref{S:emp}.

\bigskip

\Addresses


\begin{thebibliography}{}

\bibitem[Alpcan and Bauckhage, 2009]{Alpcan2009}
Alpcan, T. and Bauckhage, C. (2009).
\newblock {A distributed machine learning framework}.
\newblock {\em Proceedings of the 48h IEEE Conference on Decision and Control
  CDC held jointly with 2009 28th Chinese Control Conference}, pages
  2546--2551.

\bibitem[Attias, 1999]{attias1999inferring}
Attias, H. (1999).
\newblock Inferring parameters and structure of latent variable models by
  variational bayes.
\newblock In {\em Proceedings of the Fifteenth conference on Uncertainty in
  artificial intelligence}, pages 21--30. Morgan Kaufmann Publishers Inc.

\bibitem[Baek and Morimoto, 2012]{Baek2012}
Baek, T.~H. and Morimoto, M. (2012).
\newblock {Stay Away From Me}.
\newblock {\em Journal of Advertising}, 41(1):59--76.

\bibitem[Bakshy et~al., 2012]{Bakshy2012}
Bakshy, E., Eckles, D., Yan, R., and Rosenn, I. (2012).
\newblock {Social Influence in Social Advertising : Evidence from Field
  Experiments}.
\newblock In {\em Electronic Commerce 2012}, volume~1.

\bibitem[Benaglia et~al., 2009]{benaglia2009mixtools}
Benaglia, T., Chauveau, D., Hunter, D., and Young, D. (2009).
\newblock mixtools: An r package for analyzing finite mixture models.
\newblock {\em Journal of Statistical Software}, 32(6):1--29.

\bibitem[Bleier and Eisenbeiss, 2015]{Bleier2015}
Bleier, A. and Eisenbeiss, M. (2015).
\newblock {Personalized Online Advertising Effectiveness: The Interplay of
  What, When, and Where}.
\newblock {\em Marketing Science}, 34(5):669--688.

\bibitem[Bonfrer and Dr{\`{e}}ze, 2009]{Bonfrer2009}
Bonfrer, A. and Dr{\`{e}}ze, X. (2009).
\newblock {Real-Time Evaluation of E-mail Campaign Performance}.
\newblock {\em Marketing Science}, 28(2):251--263.

\bibitem[Boyd and Crawford, 2012]{Boyd2012}
Boyd, D. and Crawford, K. (2012).
\newblock {Critical Questions for Big Data}.
\newblock {\em Information, Communication {\&} Society}, 15(5):662----679.

\bibitem[Bucklin and Sismeiro, 2009]{Bucklin2009}
Bucklin, R.~E. and Sismeiro, C. (2009).
\newblock {Click Here for Internet Insight: Advances in Clickstream Data
  Analysis in Marketing}.
\newblock {\em Journal of Interactive Marketing}, 23(1):35--48.

\bibitem[Butler et~al., 1997]{butler1997consistency}
Butler, S.~M., Louis, T.~A., et~al. (1997).
\newblock Consistency of maximum likelihood estimators in general random
  effects models for binary data.
\newblock {\em The Annals of Statistics}, 25(1):351--377.

\bibitem[Capp{\'e} and Moulines, 2009]{cappe2009line}
Capp{\'e}, O. and Moulines, E. (2009).
\newblock On-line expectation--maximization algorithm for latent data models.
\newblock {\em Journal of the Royal Statistical Society: Series B (Statistical
  Methodology)}, 71(3):593--613.

\bibitem[Capp{\'{e}} and Moulines, 2009]{Cappe2009}
Capp{\'{e}}, O. and Moulines, E. (2009).
\newblock {On-line expectation-maximization algorithm for latent data models}.
\newblock {\em Journal of the Royal Statistical Society: Series B (Statistical
  Methodology)}, 71(3):593--613.

\bibitem[Chatterjee et~al., 2003]{Chatterjee2003}
Chatterjee, P., Hoffman, D.~L., and Novak, T.~P. (2003).
\newblock {Modeling the Clickstream: Implications for Web-Based Advertising
  Efforts}.
\newblock {\em Marketing Science}, 22(4):520--541.

\bibitem[Cheong et~al., 2011]{Cheong2011}
Cheong, Y., Leckenby, J.~D., and Eakin, T. (2011).
\newblock {Evaluating the Multivariate Beta Binomial Distribution for
  Estimating Magazine and Internet Exposure Frequency Distributions}.
\newblock {\em Journal of Advertising}, 40(1):7--24.

\bibitem[Chipman et~al., 2007]{Chipman2007}
Chipman, H.~A., George, E.~I., and McCulloch, R.~E. (2007).
\newblock {Bayesian Ensemble Learning}.
\newblock {\em Transportation Research Part B: Methodological}, 44(5):686--698.

\bibitem[Chipman et~al., 2010]{Chipman2010}
Chipman, H.~a., George, E.~I., and McCulloch, R.~E. (2010).
\newblock {BART: Bayesian additive regression trees}.
\newblock {\em The Annals of Applied Statistics}, 4(1):266--298.

\bibitem[Cho and Cheon, 2004]{Cho2004}
Cho, C.-H. and Cheon, H.~J. (2004).
\newblock {Why do people avoid advertising on the internet?}
\newblock {\em Journal of Advertising}, 33(4):89--97.

\bibitem[Chu et~al., 2007]{Chu2007}
Chu, C.-t., Kim, S.~K., Lin, Y.-a., and Ng, A.~Y. (2007).
\newblock {Map-Reduce for Machine Learning on Multicore}.
\newblock {\em Advances in neural information processing systems}, 19(23):281.

\bibitem[Dempster et~al., 1977]{Dempster1977}
Dempster, A.~P., Laird, N.~M., and Rubin, D.~B. (1977).
\newblock {Maximum Likelihood from Incomplete Data via the EM Algorithm}.
\newblock {\em Journal of the Royal Statistical Society B}, 39(1):1--22.

\bibitem[Dippold and Hruschka, 2013]{dippold2013model}
Dippold, K. and Hruschka, H. (2013).
\newblock A model of heterogeneous multicategory choice for market basket
  analysis.
\newblock {\em Review of Marketing Science}, 11(1):1--31.

\bibitem[Do and Batzoglou, 2008]{Do2008}
Do, C.~B. and Batzoglou, S. (2008).
\newblock {What is the expectation maximization algorithm?}
\newblock {\em Nature biotechnology}, 26(8):897--899.

\bibitem[Follmann and Lambert, 1991]{follmann1991identifiability}
Follmann, D.~A. and Lambert, D. (1991).
\newblock Identifiability of finite mixtures of logistic regression models.
\newblock {\em Journal of Statistical Planning and Inference}, 27(3):375--381.

\bibitem[Fr{\"u}hwirth-Schnatter, 2006]{fruhwirth2006finite}
Fr{\"u}hwirth-Schnatter, S. (2006).
\newblock {\em Finite mixture and Markov switching models}.
\newblock Springer Science \& Business Media.

\bibitem[Gaber et~al., 2005]{Gaber2005}
Gaber, M.~M., Zaslavsky, A., and Krishnaswamy, S. (2005).
\newblock {Mining data streams}.
\newblock {\em ACM SIGMOD Record}, 34(2):18.

\bibitem[Gelman, 2008]{Gelman2008}
Gelman, A. (2008).
\newblock {Objections to Bayesian statistics}.
\newblock {\em Bayesian Analysis}, 3(3):445--450.

\bibitem[Gortmaker et~al., 1994]{Gortmaker1994}
Gortmaker, S.~L., Hosmer, D.~W., and Lemeshow, S. (1994).
\newblock {Applied Logistic Regression.}

\bibitem[Graepel et~al., 2010]{graepel2010web}
Graepel, T., Candela, J.~Q., Borchert, T., and Herbrich, R. (2010).
\newblock Web-scale bayesian click-through rate prediction for sponsored search
  advertising in microsoft's bing search engine.
\newblock In {\em Proceedings of the 27th International Conference on Machine
  Learning (ICML-10)}, pages 13--20.

\bibitem[He et~al., 2014]{he2014practical}
He, X., Pan, J., Jin, O., Xu, T., Liu, B., Xu, T., Shi, Y., Atallah, A.,
  Herbrich, R., Bowers, S., et~al. (2014).
\newblock Practical lessons from predicting clicks on ads at facebook.
\newblock In {\em Proceedings of the Eighth International Workshop on Data
  Mining for Online Advertising}, pages 1--9. ACM.

\bibitem[Hoffman et~al., 2013]{hoffman2013stochastic}
Hoffman, M.~D., Blei, D.~M., Wang, C., and Paisley, J.~W. (2013).
\newblock Stochastic variational inference.
\newblock {\em Journal of Machine Learning Research}, 14(1):1303--1347.

\bibitem[Huang and Lin, 2006]{Huang2006}
Huang, C.-y. and Lin, C.-s. (2006).
\newblock {Modeling the Audience'S Banner Ad Exposure for Internet Advertising
  Planning}.
\newblock {\em Science}, 35(2):123--136.

\bibitem[Ippel et~al., 2017]{ippel2017dealing}
Ippel, L., Kaptein, M., and Vermunt, J. (2017).
\newblock Dealing with data streams: an online, row-by-row, estimation
  tutorial.
\newblock {\em Methodology: European Journal of Research Methods for the
  Behavioral and Social Sciences}.

\bibitem[Jedidi et~al., 1997]{Jedidi1997}
Jedidi, K., Jagpal, H.~S., and DeSarbo, W.~S. (1997).
\newblock {Finite-Mixture Structural Equation Models for Response-Based
  Segmentation and Unobserved Heterogeneity}.
\newblock {\em Marketing Science}, 16(1):39--59.

\bibitem[Kamakura and Russell, 1989]{kamakura1989probabilistic}
Kamakura, W.~A. and Russell, G. (1989).
\newblock A probabilistic choice model for market segmentation and elasticity
  structure.
\newblock {\em Journal of marketing research}, 26:379--390.

\bibitem[Kantas et~al., 2009]{kantas2009overview}
Kantas, N., Doucet, A., Singh, S.~S., and Maciejowski, J.~M. (2009).
\newblock An overview of sequential monte carlo methods for parameter
  estimation in general state-space models.
\newblock {\em IFAC Proceedings Volumes}, 42(10):774--785.

\bibitem[Kireyev et~al., 2015]{Kireyev2015}
Kireyev, P., Pauwels, K., and Gupta, S. (2015).
\newblock {Do display ads influence search? Attribution and dynamics in online
  advertising}.
\newblock {\em International Journal of Research in Marketing}.

\bibitem[Kruschke, 2014]{kruschke2014doing}
Kruschke, J. (2014).
\newblock {\em Doing Bayesian data analysis: A tutorial with R, JAGS, and
  Stan}.
\newblock Academic Press.

\bibitem[Kubrusly and Gravier, 1973]{kubrusly1973stochastic}
Kubrusly, C. and Gravier, J. (1973).
\newblock Stochastic approximation algorithms and applications.
\newblock In {\em 1973 IEEE Conference on Decision and Control including the
  12th Symposium on Adaptive Processes}, number~12, pages 763--766.

\bibitem[Laczniak, 2015]{Laczniak2015}
Laczniak, R.~N. (2015).
\newblock {The Journal of Advertising and the Development of Advertising
  Theory: Reflections and Directions for Future Research}.
\newblock {\em Journal of Advertising}, 44(4):429--433.

\bibitem[Liang and Klein, 2009]{liang2009online}
Liang, P. and Klein, D. (2009).
\newblock Online em for unsupervised models.
\newblock In {\em Proceedings of human language technologies: The 2009 annual
  conference of the North American chapter of the association for computational
  linguistics}, pages 611--619. Association for Computational Linguistics.

\bibitem[McFarland et~al., 2006]{McFarland2006}
McFarland, R.~G., Challagalla, G.~N., and Shervani, T.~a. (2006).
\newblock {Influence Tactics for Effective Adaptive Selling}.
\newblock {\em Journal of Marketing}, 70(4):103--117.

\bibitem[Micheaux, 2011]{Micheaux2011}
Micheaux, A.~L. (2011).
\newblock {Managing e-mail Advertising Frequency from the Consumer
  Perspective}.
\newblock {\em Journal of Advertising}, 40(4):45--66.

\bibitem[Moe and Fader, 2004]{Moe2004a}
Moe, W.~W. and Fader, P.~S. (2004).
\newblock {Capturing evolving visit behavior in clickstream data}.
\newblock {\em Journal of Interactive Marketing}, 18(1):5--19.

\bibitem[Moon, 1996]{Moon1996}
Moon, T.~K. (1996).
\newblock {The expectation-maximization algorithm}.
\newblock {\em IEEE Signal Processing Magazine}, 13:47--60.

\bibitem[Neal and Hinton, 1998]{neal1998view}
Neal, R.~M. and Hinton, G.~E. (1998).
\newblock A view of the em algorithm that justifies incremental, sparse, and
  other variants.
\newblock In {\em Learning in graphical models}, pages 355--368. Springer.

\bibitem[Opper and Winther, 1998]{Opper1998a}
Opper, M. and Winther, O. (1998).
\newblock {A Bayesian approach to on-line learning}.
\newblock {\em On-line Learning in Neural Networks, ed. D {\ldots}}.

\bibitem[Owen and Eckles, 2012]{Owen2012}
Owen, A.~B. and Eckles, D. (2012).
\newblock {Bootstrapping data arrays of arbitrary order}.
\newblock {\em The Annals of Applied Statistics}, 6(3):895--927.

\bibitem[Poggio et~al., 2011]{Poggio2011}
Poggio, T., Voinea, S., and Rosasco, L. (2011).
\newblock {Online Learning, Stability, and Stochastic Gradient Descent}.
\newblock {\em Artificial Intelligence}, 8(11):11.

\bibitem[Ranaweera, 2005]{Ranaweera2005}
Ranaweera, C. (2005).
\newblock {A model of online customer behavior during the initial transaction:
  Moderating effects of customer characteristics}.
\newblock {\em Marketing Theory}, 5(1):51--74.

\bibitem[Rojas-M{\'{e}}ndez et~al., 2009]{Rojas-Mendez2009}
Rojas-M{\'{e}}ndez, J.~I., Davies, G., and Madran, C. (2009).
\newblock {Universal differences in advertising avoidance behavior: A
  cross-cultural study}.
\newblock {\em Journal of Business Research}, 62(10):947--954.

\bibitem[Ryu et~al., 2011]{Ryu2011}
Ryu, D., Li, E., and Mallick, B.~K. (2011).
\newblock {Bayesian nonparametric regression analysis of data with random
  effects covariates from longitudinal measurements.}
\newblock {\em Biometrics}, 67(2):454--66.

\bibitem[Salloum et~al., 2016]{salloum2016big}
Salloum, S., Dautov, R., Chen, X., Peng, P.~X., and Huang, J.~Z. (2016).
\newblock Big data analytics on apache spark.
\newblock {\em International Journal of Data Science and Analytics}, pages
  1--20.

\bibitem[Schmittlein and Peterson, 1994]{schmittlein1994customer}
Schmittlein, D.~C. and Peterson, R.~A. (1994).
\newblock Customer base analysis: An industrial purchase process application.
\newblock {\em Marketing Science}, 13(1):41--67.

\bibitem[Scott et~al., 2016]{scott2016bayes}
Scott, S.~L., Blocker, A.~W., Bonassi, F.~V., Chipman, H.~A., George, E.~I.,
  and McCulloch, R.~E. (2016).
\newblock Bayes and big data: The consensus monte carlo algorithm.
\newblock {\em International Journal of Management Science and Engineering
  Management}, 11(2):78--88.

\bibitem[Seyedghorban et~al., 2015]{Seyedghorban2015}
Seyedghorban, Z., Tahernejad, H., and Matanda, M.~J. (2015).
\newblock {Reinquiry into Advertising Avoidance on the Internet: A Conceptual
  Replication and Extension}.
\newblock {\em Journal of Advertising}, pages 1--10.

\bibitem[Su and Chen, 2015]{Su2015}
Su, Q. and Chen, L. (2015).
\newblock {A method for discovering clusters of e-commerce interest patterns
  using click-stream data}.
\newblock {\em Electronic Commerce Research and Applications}, 14(1):1--13.

\bibitem[Tank et~al., 2015]{tank2015streaming}
Tank, A., Foti, N.~J., and Fox, E.~B. (2015).
\newblock Streaming variational inference for bayesian nonparametric mixture
  models.
\newblock In {\em AISTATS}.

\bibitem[Teicher, 1963]{teicher1963identifiability}
Teicher, H. (1963).
\newblock Identifiability of finite mixtures.
\newblock {\em The Annals of Mathematical Statistics}, pages 1265--1269.

\bibitem[Teicher, 1967]{teicher1967identifiability}
Teicher, H. (1967).
\newblock Identifiability of mixtures of product measures.
\newblock {\em The Annals of Mathematical Statistics}, 38(4):1300--1302.

\bibitem[Titterington et~al., 1985]{titterington1985statistical}
Titterington, D.~M., Smith, A.~F., and Makov, U.~E. (1985).
\newblock {\em Statistical analysis of finite mixture distributions}.
\newblock Wiley,.

\bibitem[Toulis et~al., 2014]{toulis2014statistical}
Toulis, P., Airoldi, E., and Rennie, J. (2014).
\newblock Statistical analysis of stochastic gradient methods for generalized
  linear models.
\newblock In {\em ICML}, pages 667--675.

\bibitem[Vakratsas et~al., 2004]{Vakratsas2004}
Vakratsas, D., Feinberg, F.~M., Bass, F.~M., and Kalyanaram, G. (2004).
\newblock {The Shape of Advertising Response Functions Revisited: A Model of
  Dynamic Probabilistic Thresholds}.
\newblock {\em Marketing Science}, 23(1):109--119.

\bibitem[Van~den Bulte and Joshi, 2007]{van2007new}
Van~den Bulte, C. and Joshi, Y.~V. (2007).
\newblock New product diffusion with influentials and imitators.
\newblock {\em Marketing Science}, 26(3):400--421.

\bibitem[Wang and Puterman, 1998]{Wang1998}
Wang, P. and Puterman, M.~L. (1998).
\newblock {Mixed Logistic Regression Models}.
\newblock {\em Journal of Agricultural, Biological, and Environmental
  Statistics}, 3(2):175.

\bibitem[Wedel and DeSarbo, 1995]{wedel1995mixture}
Wedel, M. and DeSarbo, W.~S. (1995).
\newblock A mixture likelihood approach for generalized linear models.
\newblock {\em Journal of Classification}, 12(1):21--55.

\bibitem[Wedel et~al., 1993]{wedel1993latent}
Wedel, M., DeSarbo, W.~S., Bult, J.~R., and Ramaswamy, V. (1993).
\newblock A latent class poisson regression model for heterogeneous count data.
\newblock {\em Journal of Applied Econometrics}, 8(4):397--411.

\bibitem[Wedel and Kamakura, 2012]{wedel2012market}
Wedel, M. and Kamakura, W.~A. (2012).
\newblock {\em Market segmentation: Conceptual and methodological foundations},
  volume~8.
\newblock Springer Science \& Business Media.

\bibitem[West et~al., 1997]{west1997comparative}
West, P.~M., Brockett, P.~L., and Golden, L.~L. (1997).
\newblock A comparative analysis of neural networks and statistical methods for
  predicting consumer choice.
\newblock {\em Marketing Science}, 16(4):370--391.

\bibitem[Yaveroglu and Donthu, 2008]{Yaveroglu2008}
Yaveroglu, I. and Donthu, N. (2008).
\newblock {Advertising Repetition and Placement Issues in On-Line
  Environments}.
\newblock {\em Journal of Advertising}, 37(2):31--44.

\bibitem[Yeu et~al., 2013]{Yeu2013}
Yeu, M., Yoon, H.-S., Taylor, C.~R., and Lee, D.-H. (2013).
\newblock {Are Banner Advertisements in Online Games Effective?}
\newblock {\em Journal of Advertising}, 42(May 2015):241--250.

\bibitem[Yue et~al., 2012]{Yue2012}
Yue, Y., Hong, S., and Guestrin, C. (2012).
\newblock {Hierarchical exploration for accelerating contextual bandits}.
\newblock {\em arXiv preprint arXiv:1206.6454}.

\bibitem[Zhang and Krishnamurthi, 2004]{zhang2004customizing}
Zhang, J. and Krishnamurthi, L. (2004).
\newblock Customizing promotions in online stores.
\newblock {\em Marketing Science}, 23(4):561--578.

\bibitem[Zhou et~al., 2005]{zhou2005streaming}
Zhou, J., Foster, D., Stine, R., and Ungar, L. (2005).
\newblock Streaming feature selection using alpha-investing.
\newblock In {\em Proceedings of the eleventh ACM SIGKDD international
  conference on Knowledge discovery in data mining}, pages 384--393. ACM.

\bibitem[Zinkevich et~al., 2010]{Zinkevich2010}
Zinkevich, M.~A., Smola, A., and Weimer, M. (2010).
\newblock {Parallelized Stochastic Gradient Descent}.
\newblock {\em Advances in Neural Information Processing Systems 23},
  23(6):1--9.

\end{thebibliography}
\end{document}